\documentclass[10pt,twocolumn,letterpaper]{article}

\usepackage{cvpr}
\usepackage{times}
\usepackage{epsfig}
\usepackage{amsmath}
\usepackage{amssymb}
\usepackage{multicol}
\usepackage{multirow}
\usepackage{comment}
\usepackage{graphics}

\cvprfinalcopy 

\begin{document}

\title{Semi-synthesis: A fast way to produce effective datasets for stereo matching}  

\author{
	Ju He$^{1*}$
	\;\; Enyu Zhou$^{2*}$
	\;\; Liusheng Sun$^2$
	\;\; Fei Lei$^2$
	\;\; Chenyang Liu$^2$
	\;\; Wenxiu Sun$^2$\\	
	$^1$Johns Hopkins University \;\; $^2$SenseTime Research\\	
}

\maketitle
\let\thefootnote\relax\footnotetext{*: equal contribution.}
\begin{abstract}

Stereo matching is an important problem in computer vision which has drawn tremendous research attention for decades. Recent years, data-driven methods with convolutional neural networks (CNNs) are continuously pushing stereo matching to new heights. However, data-driven methods require large amount of training data, which is not an easy task for real stereo data due to the annotation difficulties of per-pixel ground-truth disparity. Though synthetic dataset is proposed to fill the gaps of large data demand, the fine-tuning on real dataset is still needed due to the domain variances between synthetic data and real data. In this paper, we found that in synthetic datasets, close-to-real-scene texture rendering is a key factor to boost up stereo matching performance, while close-to-real-scene 3D modeling is less important. We then propose semi-synthetic, an effective and fast way to synthesize large amount of data with close-to-real-scene texture to minimize the gap between synthetic data and real data. Extensive experiments demonstrate that models trained with our proposed semi-synthetic datasets achieve significantly better performance than with general synthetic datasets, especially on real data benchmarks with limited training data.
With further fine-tuning on the real dataset, we also achieve SOTA performance on Middlebury and competitive results on KITTI and ETH3D datasets.
\end{abstract}
\begin{figure}
    \centering
    \includegraphics[width=\linewidth]{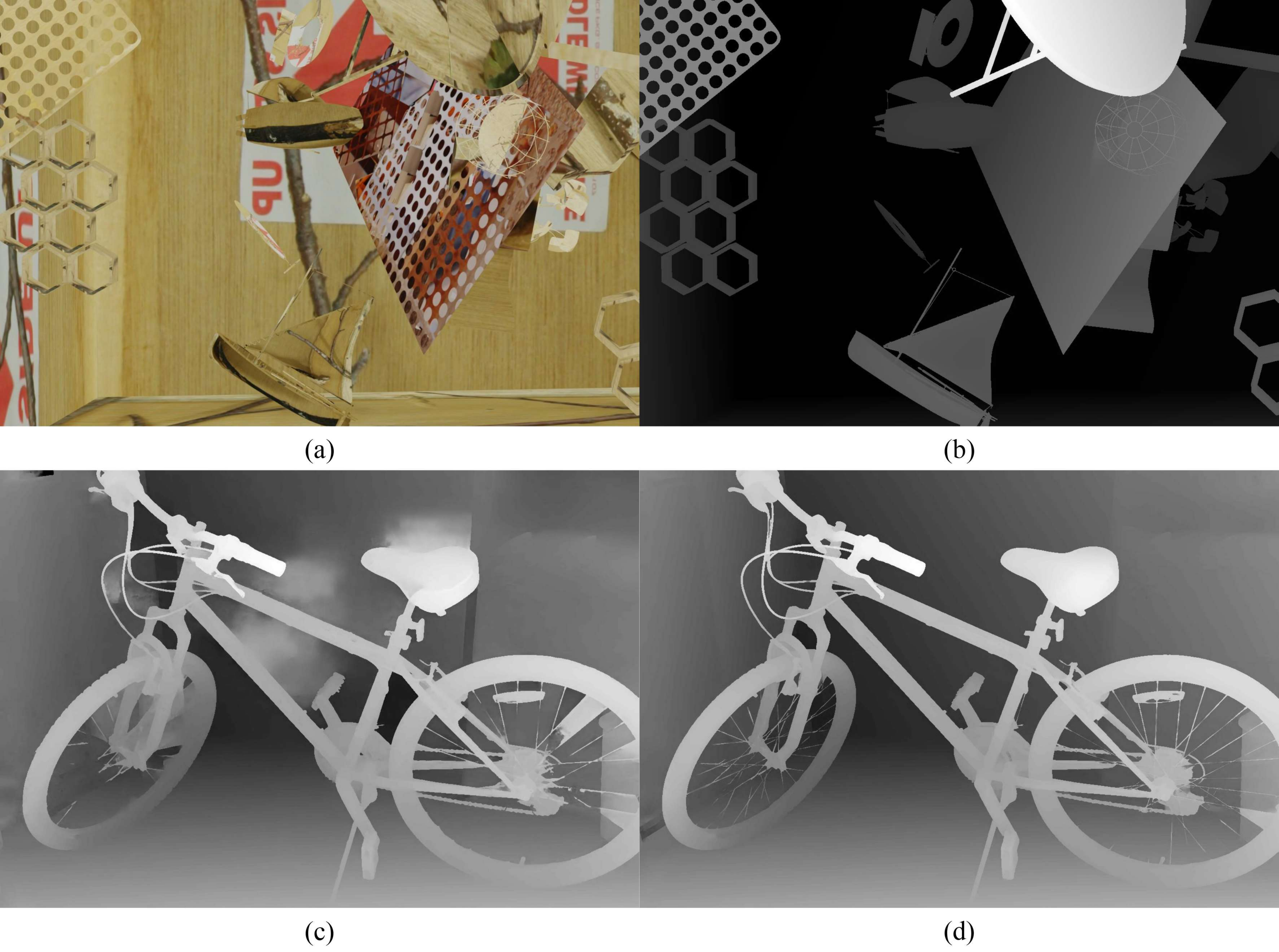}
    \caption{
    A sample pair from semi-synthetic datasets and stereo matching results on a real stereo pair from Middlebury dataset. (a) Left view of a semi-synthetic image pair. (b) Ground-truth disparity map. (c) Stereo matching result trained on SceneFlow dataset. (d) Stereo matching result trained on Semi-Synthetic datasets.
    }
    \label{fig:example}
\end{figure}

\section{Introduction}
Stereo matching is one of the most fundamental problems in computer vision. It is widely used in applications such as reconstruction \cite{izadi2011kinectfusion, geiger2011stereoscan}, robot navigation \cite{murray2000using, salmeron2015tradeoff}, and autonomous driving \cite{geiger2012we,li2019stereo}. Traditional methods solve this task by carefully hand-crafted image priors and energy functions.
Recently, with the resurgence of the deep learning, we have witnessed significant progress in this field. 
The deep learning methods define loss functions and learn complex priors but are data-thirsty, which require a large amount of training data to 
reach good performance.

There exist two kinds of datasets for stereo matching. 
One is real datasets such as Middlebury \cite{10.1007/978-3-319-11752-2_3}, KITTI \cite{menze2015object}, ETH3D \cite{schoeps2017cvpr}, and the 
other is synthetic datasets such as Sintel \cite{Butler:ECCV:2012}, Sun3D \cite{xiao2013sun3d}, SceneFlow \cite{mayer2016large}. Though they prove their effectiveness in their corresponding task, they all suffer from problems that harm their practical use. Real datasets are usually small in scale due to the amount of human labor involved in constructing the scenes and annotating ground truth information, leading to the inaccuracy of ground truth and monotonousness of scenes. On the other side, synthetic datasets suffer from lacking real textures, and they usually cost a long time to produce because of constructing real scenes and rendering. 

To overcome these shortcomings, people now usually combine these two kinds of datasets to train models. The models are firstly pre-trained on the large general synthetic datasets followed by fine-tuned on the corresponding real datasets. However, this kind of training strategy struggles when there is difficulty in collecting sufficient high-quality real data. Besides, since some key factors of synthetic datasets such as textures, illumination,
disparity distribution might be totally different from the real scenes, 
this kind of pre-training might hurt the model's performance.


Due to the limitations on the synthetic datasets, 
recent work started to research on how to tackle domain gaps between different datasets by either transferring certain dataset to others \cite{song2020adastereo, liu2020stereogan} or doing normalization to each dataset separately \cite{zhang2019domain}. 
However, there exists difficulty to transfer certain key factors such as disparity distribution and textures to a uniform space.
Their usage is also limited to existing datasets and can not be adapted to unseen scenes.

In this paper, we propose a novel and fast data synthesis method,
\emph{semi-synthesis}, to produce large-scale on-demand stereo datasets. It is called semi-synthesis because we extract real image patches from the corresponding scenes and then texture them on generated geometry shapes. 
With this simple method, we can easily control 
the key factors that affect the training performance for the datasets, such as disparity distribution, textures, 
geometry shapes, etc. Also, with the fast speed of generating an image pair, we can produce large amounts of data in a short time, which definitely facilitates the training of deep models. Extensive experiments have been conducted to prove the effectiveness of our semi-synthetic datasets. Besides the simplicity of constructing large on-demand datasets compared to traditional synthetic datasets, our semi-synthetic datasets also alleviate the problem of domain gaps by not only sampling textures from the corresponding scenes but also mimicking the desired environment factors. After only trained on our semi-synthetic datasets, models outperform those trained on SceneFlow on all real benchmarks, and even quite close to those fine-tuned on real datasets on Middlebury.

Our contributions are as follows:

\begin{enumerate}
    \item We propose a novel and fast method to produce large on-demand semi-synthetic datasets for stereo matching. This method can also be further extended to some other fields, such as optical flow.
    \item We analyze the impact of textures and scene geometry 
    of semi-synthetic datasets on the final performance.
    \item We achieve significantly better performance on stereo matching benchmarks with our semi-synthetic datasets than with general synthetic datasets. 
\end{enumerate}
\section{Related Work}

\begin{figure*}[ht]
    \centering
    \includegraphics[width=\linewidth]{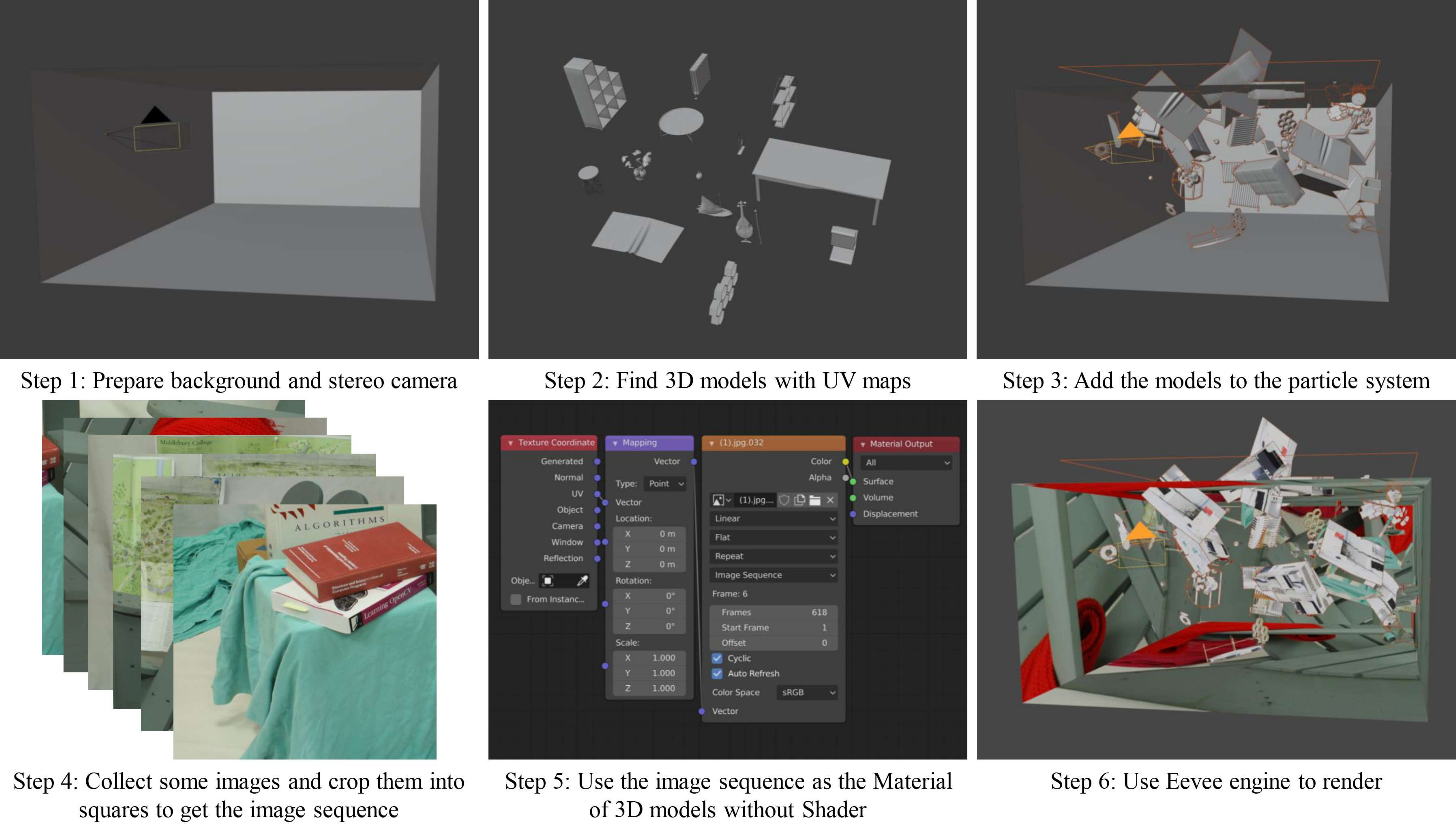}
    \caption{Diagram of the pipeline of generating semi-synthetic datasets.}
    \label{fig:model}
\end{figure*}

In this section, we briefly analyze existing stereo datasets and some recent progress on models of stereo matching.

\subsection{Stereo Datasets}

Stereo datasets can be roughly classified into two categories: 1) real datasets, 2) synthetic datasets. Real datasets are either constructed by using Time-of-Flight (ToF) or structured light. ToF is a more convenient way to collect real datasets compared to structured light. However, it struggles at the precision of ground truth since objects which are far away or opaque can not reflect any light. Another problem of ToF is that the resolution of images captured by ToF is usually small, which means tiny objects' details can not be well learned by the network. By contrast, datasets captured by structured light are much more accurate, but they are limited to indoor scenes and data collection. Thus they are usually of very small size. Synthetic datasets are usually made by first constructing background scenes and then adding some foreground objects, followed by setting stereo camera to capture the images.

We introduce four widely used datasets in detail here.

\textbf{Middlebury dataset} \cite{scharstein2002taxonomy} was one of the first datasets for stereo matching, which contains 38 real indoor scenes captured via a structured light scanner. A new version of the Middlebury dataset \cite{10.1007/978-3-319-11752-2_3} was proposed to give a more accurate annotation at a resolution of 6 Megapixels, and it contains 33 novel indoor scenes. However, due to the difficulty and high cost of constructing such accurate dense stereo datasets, they are relatively small in size, and this also yields the problem of low variability. The scenes are limited in the indoor environment with controlled light conditions.

\textbf{KITTI dataset} was first produced in 2012 and extended in 2015. It was recorded by using a laser scanner mounted on a car. While it is also a real dataset, the recorded scenes are limited on streets with a fixed height and width. Moreover, the benchmark images are of low resolution, and the laser only provides sparse ground truth information. 

\textbf{ETH3D dataset} was proposed by $Sch\ddot{o}ps$ et al. \cite{schoeps2017cvpr} in 2017, which contains real images. The ground truth of the datasets is acquired by using a laser scanner. Instead of carefully constructing scenes in a controlled laboratory environment as in Middlebury, ETH3D provides the full range of challenges of real-world photogrammetric measurements. However, it still suffers from a lack of data samples and variability.

\textbf{SceneFlow dataset} is a combined dataset of three large synthetic stereo video datasets proposed by Mayer~\cite{mayer2016large}. As a synthetic dataset, it has accurate dense disparity maps and variability in scenes. However, it still suffers problems such as the fixed disparity distribution, limiting its usage on different baselines. Besides, the cost to extend it to a specified domain can also not be ignored due to the limited scene environments.

\subsection{Models for Stereo Matching}
As introduced by the survey of Scharstein \cite{988771}, a typical stereo matching algorithm takes four steps: matching cost calculation, matching cost aggregation, disparity calculation, and disparity refinement. Traditional methods either focus on aggregating local costs according to neighborhood information \cite{6618891,4811952} or constructing a global energy function to minimize it \cite{6247827,1467526,10.1109/ICPR.2006.1033}. 

Recently, with the appearance of large-scale datasets, deep neural networks tuned for stereo matching produce SOTA performance on several stereo benchmarks. Prior work \cite{7780983,8100213,7298767} mainly focus on producing better features for traditional stereo matching algorithms with a convolutional neural network. First end-to-end network DispNet together with a large synthetic dataset ScenFlow is introduced in \cite{mayer2016large}. Then the widely used 3D cost volume is first proposed in GCNet \cite{kendall2017end}, which aims at using regression to find the optimum matching results. PSMNet \cite{chang2018pyramid} further introduced pyramid spatial pooling and 3D hourglass structures to enlarge the net's receptive field and achieve better results. 

While these methods get good results on stereo matching, they suffer from considerable use of GPU memory and time, which limits their practical use. So recent research pays more attention to fast and light methods on high-resolution images. HSM \cite{yang2019hierarchical} built a light model with a hierarchical design. AANet \cite{xu2020aanet} replaced the costly 3D convolutions with aggregation modules and achieves comparable accuracy. Cascade Cost Volume \cite{gu2020cascade} further extends the hierarchical modules method to narrow the search range of deeper stages based on the output of prior stages.

These models can achieve outstanding performance with a large number of training samples but will struggle in situations when data are insufficient. Besides, the common problem is that models trained on one specific domain can not generalize well on new domains. In this paper, we aim at solving these two problems simultaneously by generating large-scale semi-synthetic datasets in the next section.
\section{Method}
\begin{figure*}
    \centering
    \includegraphics[width=\linewidth]{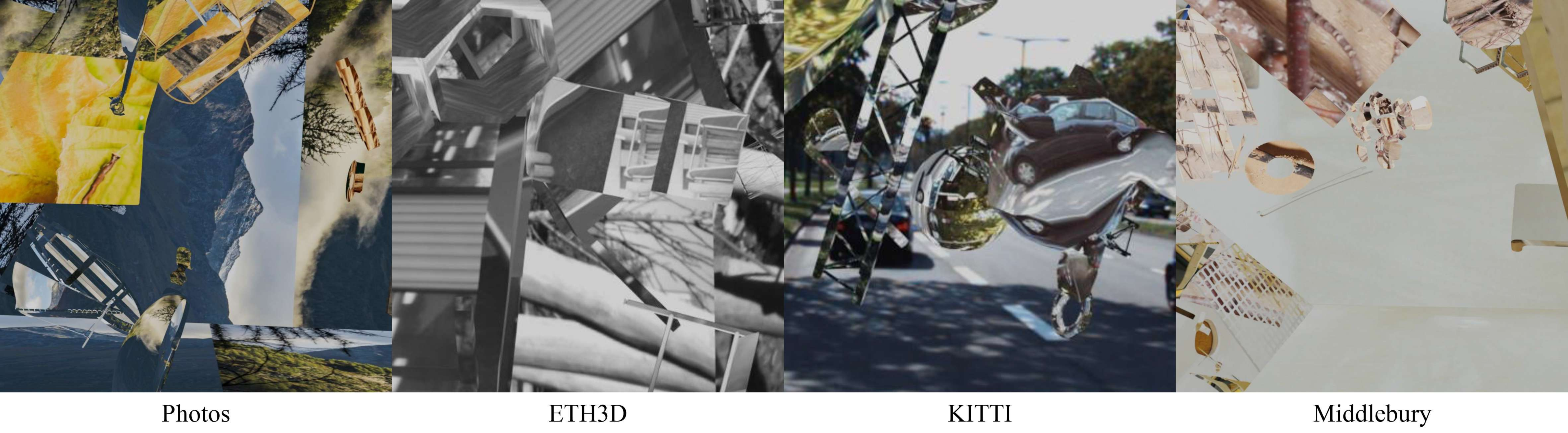}
    \caption{\textbf{Examples demonstrating the variety of our semi-synthetic datasets.} With manually setting of stereo camera's hyper-parameters, selected 3D models and image sequences, our datasets show a flexibility in terms of disparity distribution, objects, and textures.}
    \label{fig:Scenes}
\end{figure*}

We introduce our method of producing semi-synthetic datasets step by step in section \ref{sec:pipe} followed by an analysis of our main differences from other synthetic datasets at section \ref{sec:comp}.


\subsection{Pipeline}
\label{sec:pipe}
The open-source 3D software Blender 2.83a is used to generate required stereo data, including left images, right images, and the dense ground truth of disparity. As illustrated by figure \ref{fig:model}, our method mainly contains the following six steps.

\textbf{Preparing background and stereo camera.} For indoor scenes, we usually choose a cuboid as a background to simulate the walls, floors, and ceilings in the scene. For outdoor scenes, we generally choose a large faraway plane as the background whose disparity value is close to zero. The example scene shown in figure \ref{fig:model} is used to generate the semi-synthetic training datasets for Middlebury. Besides, we add the stereo camera, which is a built-in function of Blender. As the experiment conducted by \cite{mayer2018makes}, the disparity distribution of the training set will affect the results of the network. So we set the focal length, sensor size, and baseline parameters of the scene according to the testing environments to simulate the max disparity of testing scenes.

\textbf{Looking for proper 3D models with UV maps.} The desired 3D models can be divided into simple geometric primitives such as cubes and real object models like tables, which can be downloaded from the Blender kit. After acquiring these models, we delete their original materials and only keep their UV maps. Random scaling and rotation augmentation are applied for data diversity.

\textbf{Adding 3D models to the particle system.} After choosing the needed 3D models, we use Blender's particle system to make these 3D models move in the space, avoiding manually setting the motion tracks for objects. An emission source is placed in the space to emit particles (the 3D models), which is at a plane above the whole space in our figure \ref{fig:model}. And we can simulate the disparity distribution of test scene by controlling the particle quantity in different levels of depth. In this process, a sequence of frames will be captured by the stereo camera, and each frame serves as an image in the final dataset.

\textbf{Collecting textures from images.} With all settings of the 3D structure finished, we collect some pictures from the testing scenes as the textures of the 3D objects. Note that what we need is only a small set of monocular images without any other types of data, such as ground truth disparity. Collected images are further cropped into a series of squares to prevent the images from being stretched. Putting these cropped images together yields an image sequence to be exploited later.

\textbf{Texturing collected image patches to 3D models.} Each object can be randomly textured with an image sequence sampled through the build-in features of Blender, which inducing more variety in the data. Shading mechanism is not applied to our model, unlike other synthetic datasets, so the images are directly output as textures to the material. In this way, we can ensure that the imported image textures will not be disturbed by lighting so that the network can extract the real image's features. The parameter Mapping Scale in Blender is adjusted according to the image patches for avoiding over-scaling.

\textbf{Rendering the whole scene.} Finally, the Eevee engine is employed for rendering, and the depth map of the left and right cameras can be output in the Blender's Compositing module. As we do not add shader to the pipeline, our method reaches a very fast speed. Using one RTX 2080Ti, we can generate an image pair and the dense disparity ground truth at the resolution of 1500$\times$1000 in 2 seconds, saving a few minutes of rendering time. 

\begin{figure*}
    \centering
    \includegraphics[width=\linewidth]{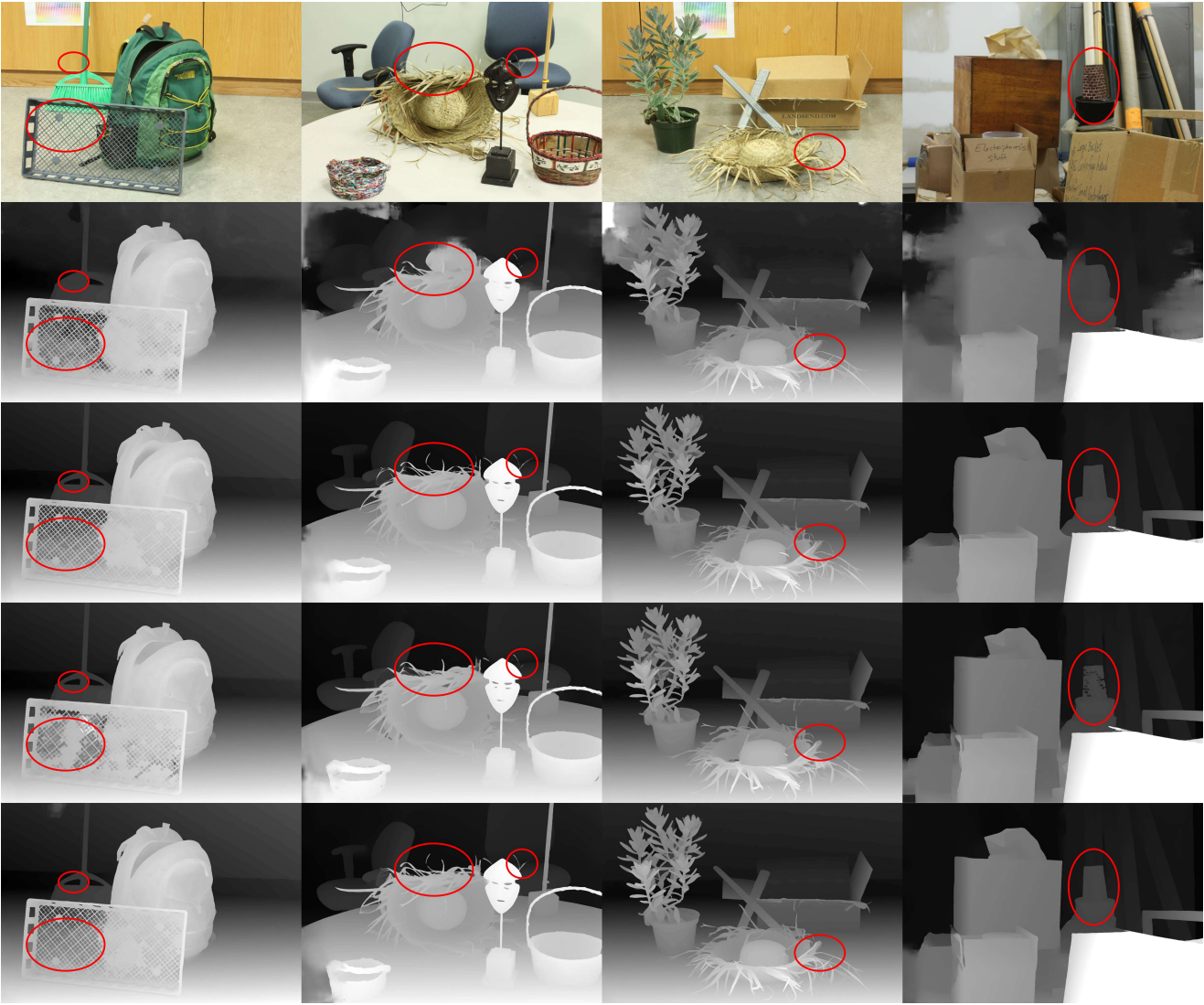}
    \caption{\textbf{Visualization of outputs of CasStereo trained on different datasets on four Middlebury images at half resolution.} Semi-Synthetic-M stands for semi-synthetic datasets with Middlebury textures. Top row: Four images in the Middlebury datasets. Second row: Output of CasStereo trained on SceneFlow. Third row: Output of CasStereo trained on Semi-Synthetic-M. Fourth row: Output of CasStereo trained on SceneFlow and Middlebury. Last row: Output of CasStereo trained on Semi-Synthetic-M and Middlebury.}
    \label{fig:show_middlebury}
\end{figure*}

\subsection{Comparison to previous synthetic datasets} 
\label{sec:comp}

Compared with previous synthetic datasets, two major differences exist in our semi-synthetic datasets, which contribute to our superior performance, and we give a detailed analysis below.

\textbf{Texture.} Compared with SceneFlow, our dataset shares the similar 3D scene design of adding foreground objects on the background, but differ on the texture. In SceneFlow, the texture of objects was chosen from combination of procedural images, fixed real images, and texture-style photographs. We argue that procedural and texture-style textures would not contribute much to the model generalization on real datasets due to different distributions. 
By contrast, we sample our textures from the images of corresponding testing scenes directly to mimic the testing situations. Also, since we abandon high-level information by texturing different image patches to objects without limitation of objects themselves, the diversity of textures becomes much larger, which benefits the training of models.

\begin{table*}
    \small
    \centering
    \caption{Results on Middlebury-v3 additional images where all pixels are evaluated. Test size stands for the test resolution of images where Q means quarter, H means half and F stands for full. Best results for each method are bolded, second best results are underlined.}
    \vspace{10pt} 
    \begin{tabular}{|c|c|c|c|c|c|c|c|}
        \hline
        Method & Test Size & Training Datasets & avgerr & rms & bad-1.0 & bad-2.0 & bad-4.0 \\
        \hline
        \multirow{4}*{PSMNet \cite{chang2018pyramid}} & \multirow{4}*{Q}
         & SceneFlow & 13.026 & 36.281 & 61.361 & 42.810 & 26.718 \\ \cline{3-8} 
         & & Semi-Synthetic-M & \underline{4.269} & \textbf{15.662} & \underline{41.270} & \underline{23.029} & \underline{12.302} \\ \cline{3-8}
         & & SceneFlow\ +\ Middlebury & 5.563 & 19.406 & 50.363 & 30.186 & 16.447 \\ \cline{3-8}
         & & Semi-Synthetic-M\ +\ Middlebury & \textbf{4.186} & \underline{15.986} & \textbf{40.623} & \textbf{22.247} & \textbf{11.995} \\ \cline{3-8}
        \hline
        \multirow{4}*{HSM \cite{yang2019hierarchical}} & \multirow{4}*{F}
         & SceneFlow & 8.806 & 26.272 & 55.267 & 36.913 & 23.122 \\ \cline{3-8}
         & & Semi-Synthetic-M & 6.402 & \underline{22.346} & 39.487 & 24.177 & 14.497 \\ \cline{3-8}
         & & SceneFlow\ +\ Middlebury & \underline{6.368} & 23.830 & \underline{38.579} & \underline{22.331} & \underline{13.849} \\ \cline{3-8}
         & & Semi-Synthetic-M\ +\ Middlebury & \textbf{5.763} & \textbf{21.418} & \textbf{31.247} & \textbf{17.529} & \textbf{10.304} \\ \cline{3-8}
        \hline
        \multirow{4}*{CasStereo \cite{gu2020cascade}} & \multirow{4}*{H}
         & SceneFlow & 17.960 & 44.488 & 50.971 & 37.948 & 28.170 \\ \cline{3-8}
         & & Semi-Synthetic-M & \textbf{4.340} & \textbf{17.936} & \textbf{25.247} & \textbf{14.090} & \textbf{8.825} \\ \cline{3-8}
         & & SceneFlow\ +\ Middlebury & 5.570 & 20.752 & 27.961 & 17.188 & 11.399 \\ \cline{3-8}
         & & Semi-Synthetic-M\ +\ Middlebury & \underline{5.419} & \underline{21.528} & \underline{25.169} & \underline{14.526} & \underline{9.331} \\ \cline{3-8}
        \hline
    \end{tabular}
    \label{tab:Middlebury}
\end{table*}

\textbf{Diversity of geometry primitives}
Previously, Watson et al. \cite{watson2020learning} also tried to use real images for making stereo matching datasets, and compared different data generation methods, such as Affine warps \cite{detone2018superpoint}, Random pasted shapes \cite{mayer2018makes} in their paper. But all the results have not been significantly improved compared to SceneFlow especially on Middlebury. We reason that it is because they lack the diversity of geometry primitives. Therefore, We add various 3D models in the generation process by texture real images on 3D models. This operation intuitively provides much more information for the network to capture.


\section{Experiments}
We conduct extensive experiments to illustrate the effectiveness of our semi-synthetic dataset. We first describe our detailed setup in section \ref{sec:setup}, including datasets, model structures and metrics. 
Then we evaluate our models on three public benchmarks with a comparison to SceneFlow. We give a detailed analysis in section \ref{sec:comp_sf} followed by ablation studies in section \ref{sec:inn} to demonstrate the effects of different settings of several key factors in our semi-synthetic datasets.

\subsection{Setup}
\label{sec:setup}

\textbf{Datasets}
We use four publicly available datasets including Middlebury-v3 \cite{10.1007/978-3-319-11752-2_3}, KITTI 2015 \cite{menze2015object}, ETH3D \cite{schoeps2017cvpr} and Sceneflow \cite{mayer2016large} plus our own semi-synthetic datasets. Middlebury-v3 contains 10 high-resolution training image pairs and 13 additional image pairs with ground truth. KITTI 2015 contains 200 low-resolution image pairs collected on streets. ETH3D consists of 27 low-resolution image pairs of both indoor and outdoor scenes. Sceneflow contains around 35k synthetic image pairs with dense ground truth.

\textbf{Implementation}
We conduct our experiments based on three network structures: PSMNet \cite{chang2018pyramid}, HSM \cite{yang2019hierarchical}, CasStereo \cite{gu2020cascade}. PSMNet is a classic work which first introduced the pyramid structure into Stereo Matching. HSM is a light network structure proposed to produce fast on-demand results under some realistic situations and can be applied to high-resolution input pairs. CasStereo is a recent work which dynamically adjusting the search range of later stages according to the results of early stages. Limited by GPU memory due to different design of network modules, size used for Middlebury datasets during training and testing are different for each network. We follow the strategies implemented by the authors while making some adaptations to train these networks. To be specific:

For PSMNet, we perform color normalization to all data. During training, images are randomly cropped to size $H=256$ and $W=512$. The max disparity is set to $192$ for KITTI, $192$ for Middlebury, and $64$ for ETH3D.

For HSM, we perform all augmentation strategies proposed by the authors to the four public datasets excluding our semi-synthetic datasets. During training, all images are resized to size $H=512$ and $W=768$. The max disparity is set to $192$ for KITTI, $768$ for Middlebury, $64$ for ETH3D.

For CasStereo, we adopt a three-stage cascade cost volume. The max disparity is set to $192$ for KITTI, $384$ for Middlebury, and $64$ for ETH3D.  Corresponding disparity hypothesis and interval are also adapted according to the max disparity.

\textbf{Metrics} 
Our evaluation is done on Middlebury-v3, KITTI 2015 and ETH3D benchmarks. We use different metrics for these three benchmarks.

For Middlebury-v3, we adopt the official metrics which contains bad-4.0 (percentage of ``bad`` pixels whose error is greater than 4.0), bad-2.0, bad-1.0 for evaluating the result under different accuracy requirement; avgerr (average absolute error in pixels) and rms (root-mean-square disparity error in pixels) count for subpixel level absolute errors. These metrics are also used for evaluating ETH3D.

For KITTI 2015, we adopt D1-all for measuring the percentage of outliers for all pixels.

\begin{figure*}
    \centering
    \includegraphics[width=\linewidth,height=11cm]{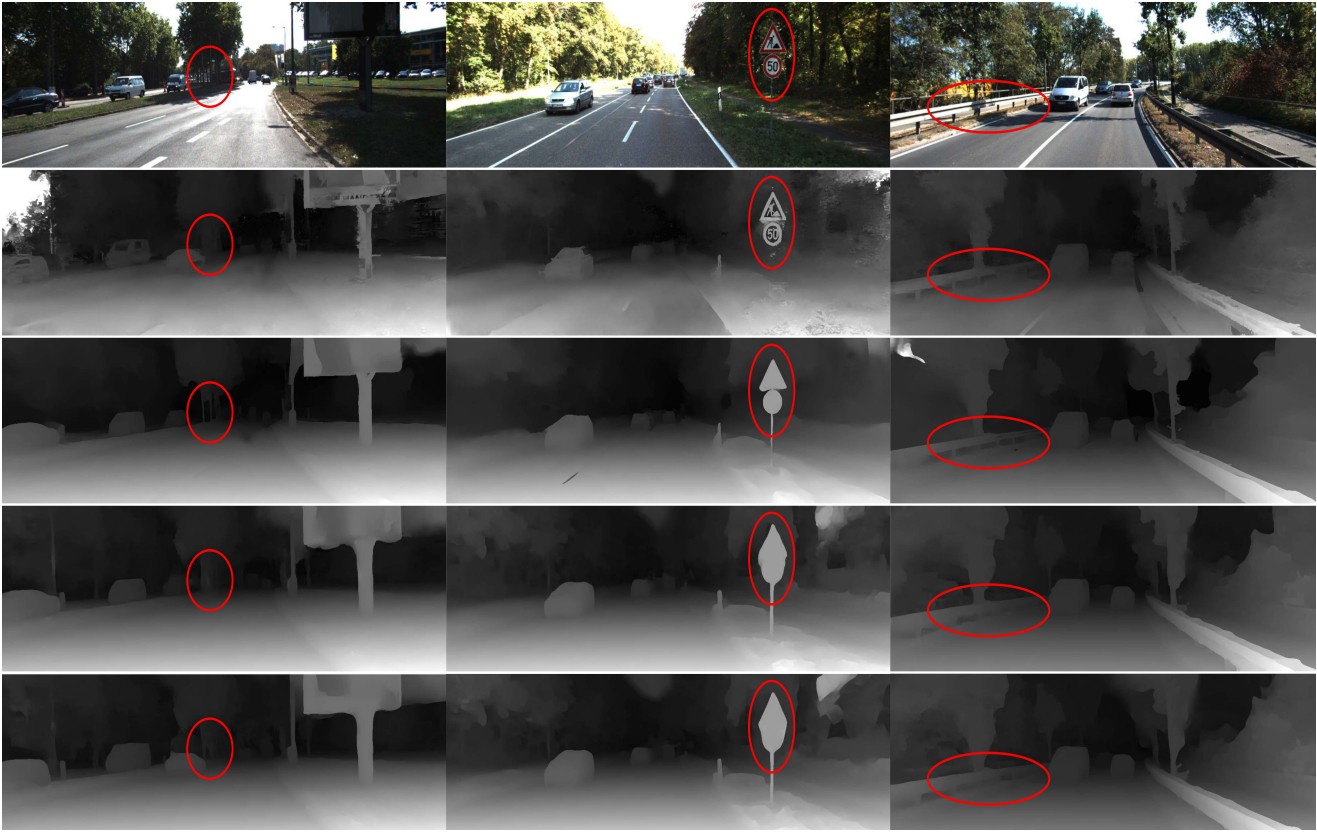}
    \caption{\textbf{Visualization of outputs of CasStereo trained on different datasets on three KITTI images.} Semi-Synthetic-K stands for semi-synthetic datasets with KITTI textures. Top row: Three images in the KITTI datasets. Second row: Output of CasStereo trained on SceneFlow. Third row: Output of CasStereo trained on Semi-Synthetic-K. Fourth row: Output of CasStereo trained on SceneFlow and KITTI. Last row: Output of CasStereo trained on Semi-Synthetic-K and KITTI.}
    \label{fig:show_kitti}
\end{figure*}

\subsection{Comparison to SceneFlow}
\label{sec:comp_sf}

In this section, We discuss several strategies for the pre-training and ﬁne-tuning of those networks. Since these benchmarks only allow one submission on the official test splits, we evaluate our experiments on the validation sets of each dataset. To be specific, for Middlebury we adopt the 13 additional pairs for validation. For KITTI 2015, we follow the split protocol in HSM \cite{yang2019hierarchical}. For ETH3D we manually sample 13 pairs from the training sets as validation sets. We mainly conduct four experiments for each network: 1) Train on SceneFlow. 2) Train on semi-synthetic datasets. 3) Train on SceneFlow and fine-tune on corresponding real datasets. 4) Train on semi-synthetic datasets and fine-tune on corresponding real datasets. For each validation dataset, we re-generate the semi-synthetic datasets with the corresponding real images as textures. We denote them with suffix, e.g. Semi-Synthetic-M means that the textures of generated semi-synthetic datasets come from Middlebury, and K stands for KITTI, E stands for ETH3D. We limit the size of semi-synthetic datasets to be comparable with SceneFlow for a fair comparison. We pre-train our models for 10 epochs and for the fine-tuning stage we set the epochs to be 30.

\textbf{Results on Middlebury.} Deep learning based methods have been struggling on Middlebury due to the insufficiency of training data and the complexity of scenes. Table \ref{tab:Middlebury} summarizes the results on Middlebury dataset. All three networks achieve much better performance when trained on semi-synthetic datasets compared to those trained on SceneFlow. The average error rate drops 8\% in average and bad-2.0 rate drops by 50\% percent. For PSMNet and CasStereo, the results are even better than those pre-trained on SceneFlow and then fine-tuned on Middlebury. Networks, pre-trained on Semi-Synthetic-M datasets and further fine-tuned on Middlebury, get little performance gain, and even negative gain for CasStereo case. This indicates that our datasets already contain sufficient information to learn the complex Middlebury scenes.

\begin{table}
    \small
    \centering
    \caption{Results on KITTI 2015 validation sets where all pixels are evaluated. Best results for each method are bolded.}
    \vspace{10pt}
    \begin{tabular}{|c|c|c|}
        \hline
        Method & Datasets & D1-all \\
        \hline
        \multirow{4}*{PSMNet \cite{chang2018pyramid}}
         & SceneFlow & 33.586 \\ \cline{2-3}
         & Semi-Synthetic-K & 9.510 \\ \cline{2-3}
         & SceneFlow\ +\ KITTI & 2.695 \\ \cline{2-3}
         & Semi-Synthetic-K\ +\ KITTI & \textbf{2.578} \\ \cline{2-3}
        \hline
        \multirow{4}*{HSM \cite{yang2019hierarchical}}
         & SceneFlow & 8.75 \\ \cline{2-3}
         & Semi-Synthetic-K & 6.73 \\ \cline{2-3}
         & SceneFlow\ +\ KITTI & 3.75 \\ \cline{2-3}
         & Semi-Synthetic-K\ +\ KITTI & \textbf{3.14} \\ \cline{2-3}
        \hline
        \multirow{4}*{CasStereo \cite{gu2020cascade}}
         & SceneFlow & 18.570 \\ \cline{2-3}
         & Semi-Synthetic-K & 4.924 \\ \cline{2-3}
         & SceneFlow\ +\ KITTI & 2.013 \\ \cline{2-3}
         & Semi-Synthetic-K\ +\ KITTI & \textbf{1.973} \\ \cline{2-3}
        \hline
    \end{tabular}
    \label{tab:KITTI}
\end{table}

\begin{table*}
    \small
    \centering
    \caption{Results on ETH3D validation sets where all pixels are evaluated. Semi-Synthetic-E stands for semi-synthetic datasets with ETH3D textures. Best results for each method are bolded.}
    \vspace{10pt}
    \begin{tabular}{|c|c|c|c|c|c|c|}
        \hline
        Method & Datasets & avgerr & rms & bad-1.0 & bad-2.0 & bad-4.0 \\
        \hline
        \multirow{4}*{PSMNet \cite{chang2018pyramid}}
         & SceneFlow & 7.042 & 15.506 & 16.460 & 9.534 & 6.878  \\ \cline{2-7}
         & Semi-Synthetic-E & 1.792 & 4.247 & 12.907 & 8.301 & 6.210 \\ \cline{2-7}
         & SceneFlow\ +\ ETH3D & 0.327 & 0.685 & 5.479 & 1.671 & 0.335  \\ \cline{2-7}
         & Semi-Synthetic-E\ +\ ETH3D & \textbf{0.296} & \textbf{0.608} & \textbf{4.455} & \textbf{1.282} & \textbf{0.305} \\ \cline{2-7}
        \hline
        \multirow{4}*{HSM \cite{yang2019hierarchical}}
         & SceneFlow & 3.282 & 8.420 & 25.020 & 11.594 & 6.564 \\ \cline{2-7}
         & Semi-Synthetic-E & 1.294 & 3.178 & 10.205 & 6.241 & 4.205 \\ \cline{2-7}
         & SceneFlow\ +\ ETH3D & 0.411 & 0.812 & 5.913 & 1.892 & 0.496 \\ \cline{2-7}
         & Semi-Synthetic-E\ +\ ETH3D & \textbf{0.360} & \textbf{0.642} & \textbf{5.688} & \textbf{1.703} & \textbf{0.376} \\ \cline{2-7}
        \hline
        \multirow{4}*{CasStereo \cite{gu2020cascade}}
         & SceneFlow & 0.521 & 1.541 & 9.033 & 3.830 & 1.836 \\ \cline{2-7}
         & Semi-Synthetic-E & 0.429 & 1.283 & 6.388 & 3.211 & 1.565 \\ \cline{2-7}
         & SceneFlow\ +\ ETH3D & 0.281 & 0.563 & 3.988 & 0.985 & 0.295 \\ \cline{2-7}
         & Semi-Synthetic-E\ +\ ETH3D & \textbf{0.257} & \textbf{0.540} & \textbf{3.333} & \textbf{0.826} & \textbf{0.261} \\ \cline{2-7}
        \hline
    \end{tabular}
    \vspace{5pt}
    \label{tab:ETH3D}
\end{table*}

\begin{table*}
    \small
    \centering
    \caption{Ablation study of texture on Middlebury.}
    \vspace{10pt}
    \begin{tabular}{|c|c|c|c|c|c|c|c|}
        \hline
        Method & Test Size & Texture & avgerr & rms & bad-1.0 & bad-2.0 & bad-4.0 \\
        \hline
        \multirow{3}*{CasStereo \cite{gu2020cascade}} & \multirow{3}*{H}
         & SceneFlow & 17.960 & 44.488 & 50.971 & 37.948 & 28.170 \\ \cline{3-8}
         & & Photos & 4.892 & 20.060 & 28.564 & 17.175 & 11.107 \\ \cline{3-8}
         & & Middlebury & \textbf{4.340} & \textbf{17.936} & \textbf{25.247} & \textbf{14.090} & \textbf{8.825} \\ \cline{3-8}
        \hline
    \end{tabular}
    \vspace{5pt}
    \label{tab:Texture}
\end{table*}

\begin{table*}[!h]
    \small
    \centering
    \caption{Ablation study of geometry shape complexity on Middlebury.}
    \vspace{10pt}
    \begin{tabular}{|c|c|c|c|c|c|c|c|}
        \hline
        Method & Test Size & Datasets & avgerr & rms & bad-1.0 & bad-2.0 & bad-4.0 \\
        \hline
        \multirow{2}*{CasStereo \cite{gu2020cascade}} & \multirow{2}*{H}
         & Simple & 5.090 & 20.419 & 28.768 & 16.208 & 9.794 \\ \cline{3-8}
         & & Complex  & \textbf{4.340} & \textbf{17.936} & \textbf{25.247} & \textbf{14.090} & \textbf{8.825} \\ \cline{3-8}
        \hline
    \end{tabular}
    \vspace{5pt}
    \label{tab:Complexity}
\end{table*}

\textbf{Results on KITTI.} KITTI is a relatively easier dataset for stereo matching since the scenes are comparable simple and consistent and there exists adequate frames. Deep models can get very good results on it after fine-tuned for many epochs, e.g. the final submission model of PSMNet on KITTI benchmark was fine-tuned for 1000 epochs. However, in this way, the models will be overfitting to the KITTI dataset and thus the performance will decrease a lot on other datasets. We only fine-tune the models on KITTI 2015 training sets for 30 epochs and compare the results. As shown in table \ref{tab:KITTI}, networks pre-trained on semi-synthetic-K improve by a lot margin compared to those pre-trained on SceneFlow. And the performance is still better than the SceneFlow ones after fine-tuned which demonstrates the superiority of our semi-synthetic datasets.

\textbf{Results on ETH3D.} ETH3D has a sparse ground truth collected by laser scanner, there exists some regions without ground truth that can not be learned sufficiently and the details of objects edge are blurred. This phenomenon can be alleviated by our semi-synthetic datasets with dense ground truth and finer details. Table \ref{tab:ETH3D} summarizes the results on ETH3D dataset. Models trained on semi-synthetic datasets outperform those trained on SceneFlow especially when not further fine-tuned on ETH3D.

\subsection{Ablation Study}
\label{sec:inn}

In this section, we conduct several ablation studies on our semi-synthetic datasets to analyze the impact of several factors, e.g. the source of textures. Our experiment in this section is based on CasStereo network for Middlebury additional images of half-size resolution. Models are only trained on semi-synthetic datasets with the corresponding factor varying and all other settings fixed. 

\textbf{Texture.} Table \ref{tab:Texture} summarizes the results on Middlebury dataset with three different texture sources. SceneFlow represents traditional synthetic datasets; photos refers to randomly downloaded images; Middlebury stands for training sets of Middlebury dataset. The texture similarity between the training and test data is $Middlebury > Photos > SceneFlow$. The results indicate that the performance becomes better with the increment of similarity between training and testing data textures, especially when the source of textures change from synthesis to reality.

\textbf{Geometry shape diversity.} Table \ref{tab:Complexity} summarizes Middlebury's performance with varying diversity of geometry shapes. Two typical settings are compared, where Simple refers to a set of simple geometric bodies, such as cuboid, cone, cylinder and etc, while Complex denotes the more complex 3D models downloaded from the Internet as shown in Figure \ref{fig:model} Step 2, which is more diverse than the former one. It is shown that a more diverse scene helps to train the network. 

\section{Conclusion}

In this work, we mainly study the problem of generating effective datasets for stereo matching. Existing real datasets are usually small in size, which hinders the training of deep models. On the other hand, general synthetic datasets suffer from lacking real textures and they do not coincide with real testing scenes on several environmental factors. We propose to solve this by using a novel and fast method to produce large-scale on-demand semi-synthetic datasets. Our extensive experiments demonstrate the effectiveness of semi-synthetic datasets on three widely used stereo benchmarks of real scenes. For future work, we aim at continuously analyzing the key factors of a good semi-synthetic dataset and extending this method to other related fields such as optical flow. 

{\small
\bibliographystyle{ieee_fullname}
\bibliography{egbib}
}

\end{document}